\documentclass[letterpaper, 10 pt, conference]{ieeeconf}  

\IEEEoverridecommandlockouts                              

\usepackage{graphics} 
\usepackage{epsfig} 
\usepackage{mathptmx} 
\usepackage{times} 
\usepackage{amsmath} 
\usepackage{amssymb}  
\usepackage{color}
\usepackage{siunitx}
\usepackage{soul}

\usepackage{cite}
\usepackage{algorithmic}
\usepackage{graphicx}
\usepackage{textcomp}

\usepackage{times}
\usepackage{epsfig}
\usepackage{amsmath}
\usepackage{amssymb}
\usepackage{slashbox}
\usepackage{subfig}
\usepackage{caption}
\usepackage{lmodern}
\usepackage{booktabs}
\usepackage{etoolbox}
\usepackage{lipsum, array, booktabs, caption}

\usepackage{filecontents}

\newcommand{\ie}{{\em i.e.}}
\newcommand{\eg}{{\em e.g.}}
\newcommand{\etal}{{\em et al.}}

\newcommand{\Fig}[1]{Fig. \ref{fig:#1}}

\newcommand{\Sect}[1]{Sect. \ref{sec:#1}}
\newcommand{\Tab}[1]{Table \ref{tab:#1}}

\usepackage{graphicx}
\graphicspath{{./figures/}}
\DeclareGraphicsExtensions{.png , .jpg, .eps}

\usepackage[usernames]{xcolor}

\usepackage{verbatim} 

\title{\LARGE \bf
Monocular Depth Estimation by Learning from Heterogeneous Datasets
}

\author{Akhil Gurram$^{1, 2}$, Onay Urfalioglu$^{2}$, Ibrahim Halfaoui$^{2}$, Fahd Bouzaraa$^{2}$ and Antonio M. L\'opez$^{1}$ 
\thanks{$^{1}$Akhil Gurram and Antonio M. L\'opez are with The Computer Vision Center (CVC) and with the Dpt. of Computer Science, Universitat Aut\`onoma de Barcelona (UAB) %
	   {\tt\small akhil.gurram@e-campus.uab.cat ; antonio@cvc.uab.es} } %
\thanks{$^{2}$Akhil Gurram, Onay Urfalioglu, Ibrahim halfaoui and Fahd Bouzaraa are with Huawei European Research Center, 80992 M{\"u}nchen, Germany. %
       {\tt\small \{akhil.gurram | onay.urfalioglu | ibrahim.halfaoui | bouzaraa.fahd\}~@huawei.com} } %
}

\begin{document}

\maketitle
\thispagestyle{empty}
\pagestyle{empty}

\begin{abstract}
Depth estimation provides essential information to perform autonomous driving and driver assistance. Especially, Monocular Depth Estimation is interesting from a practical point of view, since using a single camera is cheaper than many other options and avoids the need for continuous calibration strategies as required by stereo-vision approaches. State-of-the-art methods for Monocular Depth Estimation are based on Convolutional Neural Networks (CNNs). A promising line of work consists of introducing additional semantic information about the traffic scene when training CNNs for depth estimation. In practice, this means that the depth data used for CNN training is complemented with images having pixel-wise semantic labels, which usually are difficult to annotate (e.g. crowded urban images). Moreover, so far it is common practice to assume that the same raw training data is associated with both types of ground truth, i.e., depth and semantic labels. The main contribution of this paper is to show that this hard constraint can be circumvented, i.e., that we can train CNNs for depth estimation by leveraging the depth and semantic information coming from heterogeneous datasets. In order to illustrate the benefits of our approach, we combine KITTI depth and Cityscapes semantic segmentation datasets, outperforming state-of-the-art results on Monocular Depth Estimation.
\end{abstract}

\section{Introduction}

Depth estimation provides essential information at all levels of driving assistance and automation. Active sensors such a RADAR and LIDAR provide sparse depth information. Post-processing techniques can be used to obtain dense depth information from such sparse data~\cite{Premebida:2014}. In practice, active sensors are calibrated with cameras to perform scene understanding based on depth and semantic information. Image-based object detection \cite{Yang:2016}, classification \cite{Zhu:2016}, and segmentation \cite{He:2017}, as well as pixel-wise semantic segmentation \cite{Zhao:2017} are key technologies providing such semantic information. 

Since a camera sensor is often involved in driving automation, obtaining depth directly from it is an appealing approach and so has been a traditional topic from the very beginning of ADAS\footnote{ADAS: Advanced driver-assistance systems} development. Vision-based depth estimation approaches can be broadly divided in stereoscopic and monocular camera based settings. The former includes attempts to mimic binocular human vision. Nowadays, there are robust methods for dense depth estimation based on stereo vision \cite{Hirschmuller:2008}, able to run in real-time \cite{Hernandez:2016}. However, due to operational characteristics, the mounting and installation properties, the stereo camera setup can loose calibration. This can compromise depth accuracy and may require to apply on-the-fly calibration procedures \cite{Dang:2009, Rehder:2017}. 

\begin{figure}[t!]
\centering
\includegraphics[width=\columnwidth]{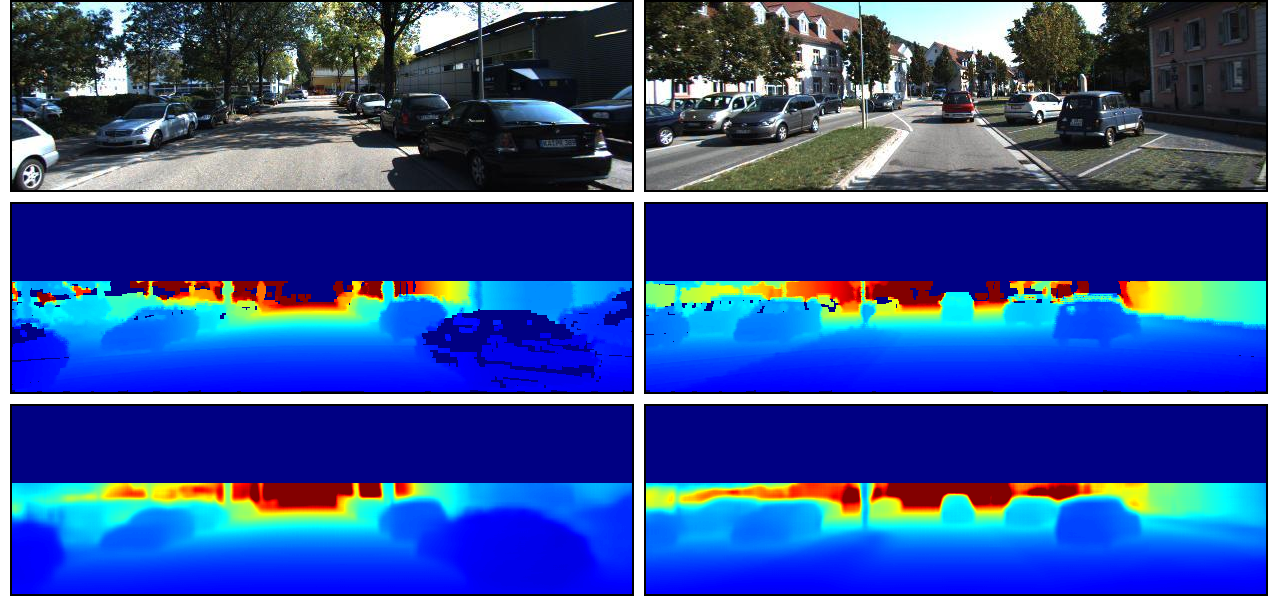}
\caption{Top to bottom: RGB KITTI images; their depth ground truth (LIDAR); our monocular depth estimation.}
\label{fig:mainresults}
\end{figure}

On the other hand, monocular depth estimation would solve the calibration problem. Compared to the stereo setting, one disadvantage is the lack of the scale information, since stereo cameras allow for direct estimation of the scale by triangulation. Though, from a theoretical point of view, there are other depth cues such as occlusion and semantic object size information which are successfully determined by the human visual system \cite{Cutting:1995}. These cues can be exploited in monocular vision for estimating scale and distances to traffic participants. Hence, monocular depth estimation can indeed support detection and tracking algorithms \cite{Ponsa:2005, Hoiem:2008, Cheda:2012}. Dense monocular depth estimation is also of great interest since higher level 3D scene representations, such as the well-known Stixels \cite{Badino:2009, Hernandez:2017} or semantic Stixels \cite{Schneider:2016}, can be computed on top. Attempts to address dense monocular depth estimation can be found based on either super-pixels \cite{Saxena:2009} or pixel-wise semantic segmentation \cite{Liu:2010}; but in both cases relying on hand-crafted features and applied to photos mainly dominated by static traffic scenes.

State-of-the-art approaches to monocular dense depth estimation rely on CNNs \cite{Cao:2017, Fu:2017, Godard:2017, Kuznietsov:2017}. Recent work has shown \cite{Mousavian:2016, Jafari:2017} that combining depth and pixel-wise semantic segmentation in the training dataset can improve the accuracy. These methods require that each training image has per-pixel association with depth and semantic class ground truth labels, e.g., obtained from a RGB-D camera. Unfortunately, creating such datasets imposes a lot of effort, especially for outdoor scenarios. Currently, no such dataset is publicly available for autonomous driving scenarios. Instead, there are several popular datasets, such as KITTI containing depth \cite{Geiger:2013} and Cityscapes \cite{Cordts:2016} containing semantic segmentation labels. However, none of them contains both depth and semantic ground truth for the same set of RGB images. 


Depth ground truth usually relies on a $360^\circ$ LIDAR calibrated with a camera system, and the manual annotation of pixel-wise semantic classes is quite time consuming ({\eg} 60-90 minutes per image). Furthermore, in future systems, the $360^\circ$ LIDAR may be replaced by four-planes LIDARs having a higher degree of sparsity of depth cues, which makes accurate monocular depth estimation even more relevant.

Accordingly, in this paper we propose a new method to train CNNs for monocular depth estimation by leveraging depth and semantic information from multiple \emph{heterogeneous} datasets. In other words, the training process can benefit from a dataset containing only depth ground truth for a set of images, together with a different dataset that only contains pixel-wise semantic ground truth (for a different set of images). In \Sect{relatedwork} we review the state-of-the-art on monocular dense depth estimation, whereas in \Sect{method} we describe our proposed method in more detail. \Sect{experiments} shows quantitative results for the KITTI dataset, and qualitative results for KITTI and Cityscapes datasets. In particular, by combining KITTI depth and Cityscapes semantic segmentation datasets, we show that the proposed approach can outperform the state-of-the-art in KITTI (see \Fig{mainresults}). Finally, in \Sect{conclusion} we summarize the presented work and draw possible future directions.

\section{Related Work}
\label{sec:relatedwork}

First attempts to perform monocular dense depth estimation relied on hand-crafted features \cite{Liu:2010, Ladicky:2014}. However, as in many other Computer Vision tasks, CNN-based approaches are currently dominating the state-of-the-art, and so our approach falls into this category too. 

Eigen {\etal} \cite{Eigen:2014} proposed a CNN for coarse-to-fine depth estimation. Liu {\etal} \cite{Liu:2016} presented a network architecture with a CRF-based loss layer which allows end-to-end training. Laina {\etal} \cite{Laina:2016} developed an encoder-decoder CNN with a reverse Huber loss layer. Cao {\etal} \cite{Cao:2017} discretized the ground-truth depth into several bins (classes) for training a FCN-residual network that predicts these classes pixel-wise; which is followed by a CRF post-processing enforcing local depth coherence. Fu  {\etal} \cite{Fu:2017} proposed a hybrid model between classification and regression to predict high-resolution discrete depth maps and low-resolution continuous depth maps simultaneously. Overall, we share with these methods the use of CNNs as well as tackling the problem as a combination of classification and regression when using depth ground truth, but our method also leverages pixel-wise semantic segmentation ground truth during training (not needed in testing) with the aim of producing a more accurate model, which will be confirmed in \Sect{experiments}.

There are previous methods using depth and semantics during training. The motivation behind is the importance of object borders and, to some extent, object-wise consistency in both tasks (depth estimation and semantic segmentation). Arsalan {\etal} \cite{Mousavian:2016} presented a CNN consisting of two separated branches, each one responsible for minimizing corresponding semantic segmentation and depth estimation losses during training. Jafari {\etal} \cite{Jafari:2017} introduced a CNN that fuses state-of-the-art results for depth estimation and semantic labeling by balancing the cross-modality influences between the two cues. Both \cite{Mousavian:2016} and \cite{Jafari:2017} assume that for each training RGB image it is available pixel-wise depth and semantic class ground truths. Training and testing is performed in indoor scenarios, where a RGB-D integrated sensor is used (neither valid for outdoor scenarios nor for distances beyond 5 meters). In fact, the lack of publicly available datasets with such joint ground truths has limited the application of these methods outdoors. In contrast, a key of our proposal is the ability of leveraging disjoint depth and semantic ground truths from different datasets, which has allowed us to address driving scenarios. 

The works introduced so far rely on deep supervised training, thus eventually requiring abundant high quality depth ground truth. Therefore, alternative unsupervised and semi-supervised approaches have been also proposed, which rely on stereo image pairs for training a disparity estimator instead of a depth one. However, at testing time the estimation is done from monocular images. Garg {\etal} \cite{Garg:2016} trained a CNN where the loss function describes the photometric reconstruction error between a rectified stereo pair of images. Godard {\etal} \cite{Godard:2017} used a more complex loss function with additional terms for smoothing and enforcing left-right consistency to improve convergence during CNN training. Kuznietsov {\etal} \cite{Kuznietsov:2017} proposed a semi-supervised approach to estimate inverse depth maps from the CNN by combining an appearance matching loss similar to the one suggested in \cite{Godard:2017} and a supervised objective function using sparse ground truth depth coming from LIDAR. This additional supervision helps to improve accuracy estimation over \cite{Godard:2017}. All these approaches have been challenged with driving data and are the current state-of-the-art.

Note that autonomous driving is pushing forward 3D mapping, where $360^\circ$ LIDAR sensing plays a key role. Thus, calibrated depth and RGB data are regularly generated. Therefore, although, unsupervised and semi-supervised approaches are appealing, at the moment we have decided to assume that depth ground truth is available; focusing on incorporating RGB images with pixel-wise class ground truth during training. Overall, our method outperforms the state-of-the-art (\Sect{experiments} ).

\section{Proposed Approach for Monocular Depth Estimation}
\label{sec:method}

\subsection{Overall training strategy}

\begin{figure}[t!]
	\centering
	\includegraphics[width=0.475\columnwidth]{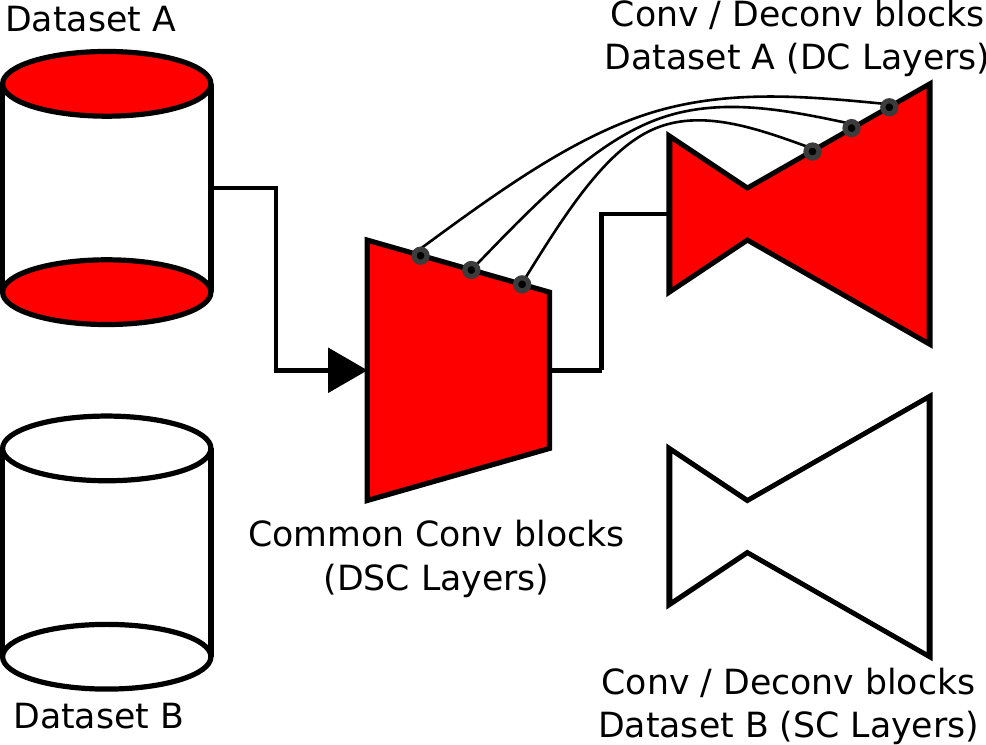}\hspace{0.04\columnwidth}\includegraphics[width=0.475\columnwidth]{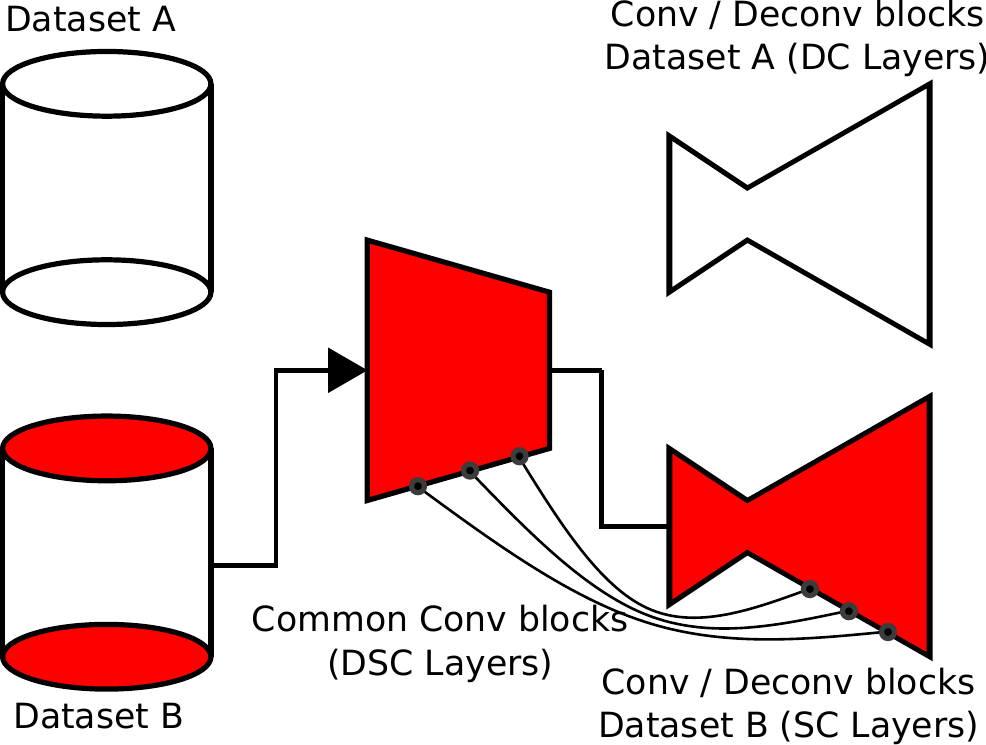}
	\caption{Phase one: conditional backward passes (see main text). We also use skip connections linking convolutional and deconvolutional layers with equal spatial sizes.}
	\label{fig:Net_Arch}
\end{figure}

\begin{figure}[t!]
	\centering
	\includegraphics[width=0.8\columnwidth]{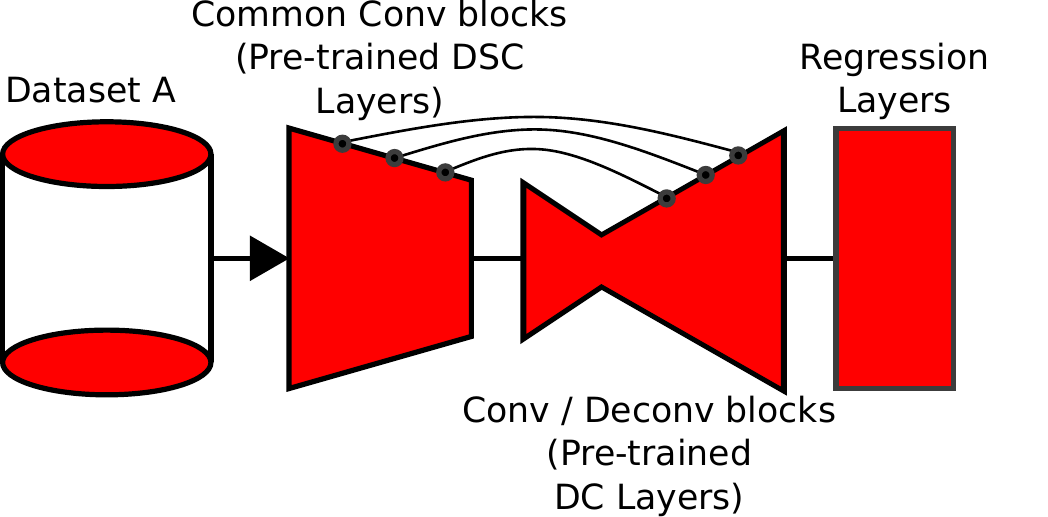}
	\caption{Phase two: the pre-trained (DSC+DC) network is augmented by regression layers for fine-tuning, resulting in the (DSC-DRN) network for depth estimation.}
	\label{fig:Net_Arch-regression}
\end{figure}

As we have mentioned, in contrast to previous works using depth and semantic information, we propose to leverage heterogeneous datasets to train a single CNN for depth estimation; {\ie} training can rely on one dataset having only depth ground truth, and a different dataset having only pixel-wise semantic labels. To achieve this, we divide the training process in two phases. In the first phase, we use a multi-task learning approach for pixel-wise depth and semantic CNN-based classification (\Fig{Net_Arch}). This means that at this stage depth is discretized, a task that has been shown to be useful for supporting instance segmentation \cite{Uhrig:2016}. In the second phase, we focus on depth estimation. In particular, we add CNN layers that perform regression taking the depth classification layers as input (\Fig{Net_Arch-regression}). 

Multi-task learning has been shown to improve the performance of different visual models ({\eg} combining semantic segmentation and surface normal prediction tasks in indoor scenarios; combining object detection and attribute prediction in PASCAL VOC images) \cite{Misra:2016}. We use a network architecture consisting of one common sub-net followed by two additional sub-net branches. We denote the layers in the common sub-net as DSC (depth-semantic classification) layers, the depth specific sub-net as DC layers and the semantic segmentation specific sub-net as SC layers. At training time we apply a conditional calculation of gradients during back-propagation, which we call \emph{conditional flow}. More specifically, the common sub-net is always active, but the origin of each data sample determines which specific sub-net branch is also active during back-propagation (\Fig{Net_Arch}). We alternate batches of depth and semantic ground truth samples.

Phase one mainly aims at obtaining a depth model (DSC+DC). Incorporating semantic information provides cues to preserve depth ordering and per object depth coherence (DSC+DS). Then, phase two uses the pre-trained depth model (DSC+DC), which we further extend by regression layers to obtain a depth estimator, denoted by DSC-DRN (\Fig{Net_Arch-regression}). We use standard losses for classification and regression tasks, {\ie} cross-entropy and L1 losses respectively.

\begin{figure*}
	\centering
	\includegraphics[width=\textwidth]{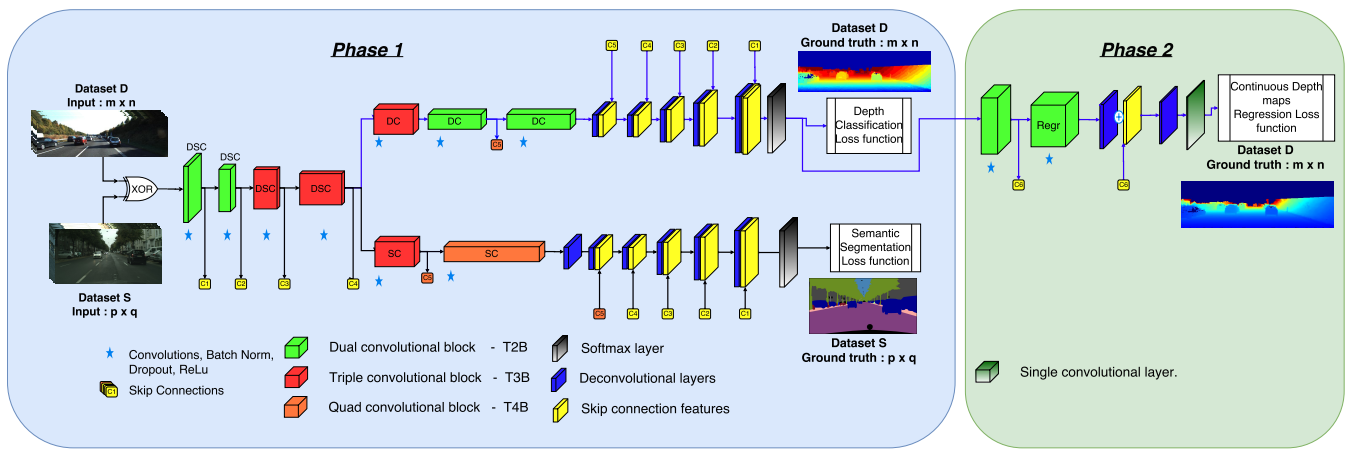}
	\caption{Details of our CNN architecture at training time.}
	\label{fig:Net_arch_}
\end{figure*}

\subsection{Network Architecture}
 
Our CNN architecture is inspired by the FCNDROPOUT of Ros \cite{Ros:2016}, which follows a convolution-deconvolution scheme. \Fig{Net_arch_} details our overall CNN architecture. First, we define a basic set of four consecutive layers consisting of Convolution, Batch Normalization, Dropout and ReLu. We build \emph{convolutional blocks} (ConvBlk) based on this basic set. There are blocks containing a varying number of sets, starting from two to four sets. The different sets of a block are chained and put in a pipeline. Each block is followed by an average-pooling layer. \emph{Deconvolutional blocks} (DeconvBlk) are based on one deconvolution layer together with skip connection features to provide more scope to the  learning process. Note that to achieve better localization accuracy these features originate from the common layers (DSC) and are bypassed to both the depth classification (DC) branch and the semantic segmentation branch (SC). In the same way we introduce skip connections between the ConvBlk and DeconvBlk of the added regression layers. 

At phase 1, the network comprises 9 ConvBlk and 11 DeconvBlk elements. At phase 2, only the depth-related layers are active. By adding 2 ConvBlk with 2 DeconvBlk elements to the (DSC+DC) branch we obtain the (DSC-DRN) network. Here, the weights of the (DSC+DC)-network part are initialized from phase 1. Note that at testing time only the depth estimation network (DSC-DRN) is required, consisting of 9 ConvBlk and 7 DeconvBlk elements.

\section{Experimental Results}
\label{sec:experiments}

\subsection{Datasets}

We evaluate our approach on KITTI dataset \cite{Geiger:2013}, following the commonly used Eigen {\etal} \cite{Eigen:2014} split for depth estimation. It consists of 22,600 training images and 697 testing images, {\ie} RGB images with associated LIDAR data. To generate dense depth ground truth for each RGB image we follow Premebida {\etal} \cite{Premebida:2014}. We use half down-sampled images, {\ie} $188 \times 620$ pixels, for training and testing. Moreover, we use 2,975 images from Cityscapes dataset \cite{Cordts:2016} with per-pixel semantic labels. 


\subsection{Implementation Details}
We implement and train our CNN using MatConvNet \cite{Vedaldi:2015}, which we modified to include the conditional flow back-propagation. We use a batch size of 10 and 5 images for depth and semantic branches, respectively. We use ADAM, with a momentum of $0.9$ and weight decay of $0.0003$. The ADAM parameters are pre-selected as $\alpha=0.001$, $\beta_1=0.9$ and $\beta_2=0.999$.  Smoothing is applied via $L_{0}$ gradient minimization  \cite{Xu:2011} as pre-processing for RGB images, with $\lambda=0.0213$ and $\kappa=8$. We include data augmentation consisting of small image rotations, horizontal flip, blur, contrast changes, as well as salt \& pepper, Gaussian, Poisson and speckle noises. For performing depth classification we have followed a linear binning of 24 levels on the range [1,80)m.

\subsection{Results}


We compare our approach to supervised methods such as Liu {\etal} \cite{Liu:2016} and Cao {\etal} \cite{Cao:2017}, unsupervised methods such as Garg {\etal} \cite{Garg:2016} and Godard {\etal} \cite{Godard:2017}, and semi-supervised method Kuznietsov {\etal} \cite{Kuznietsov:2017}. Liu {\etal}, Cao {\etal} and Kuznietsov {\etal} did not release their trained model, but they reported their results on the Eigen {\etal} split as us. Garg {\etal} and Godard {\etal} provide a Caffe model and a Tensorflow model respectively, trained on our same split (Eigen {\etal}'s KITTI split comes from stereo pairs). We have followed the author's instructions to run the models for estimating disparity and computing final depth by using the camera parameters of the KITTI stereo rig (focal length and baseline). In addition to the KITTI data, Godard {\etal} also added 22,973 stereo images coming from Cityscapes; while we use 2,975 from Cityscapes semantic segmentation challenge (19 classes). Quantitative results are shown in \Tab{SOTA_KITTI} for two different distance ranges, namely [1,50]m (cap 50m) and [1,80]m (cap 80m). As for the previous works, we follow the metrics proposed by Eigen {\etal}. Note how our method outperforms the state-of-the-art models in all metrics but one (being second best), in the two considered distance ranges. 

In \Tab{SOTA_KITTI} we also assess different aspects of our model. In particular, we compare our depth estimation results with (DSC-DRN) and without the support of the semantic segmentation task. In the latter case, we distinguish two scenarios. For the first one, which we denote as DC-DRN, we discard the SC subnet from the $1^{st}$ phase so that we first train the depth classifier and later add the regression layers for retraining the network. In the second scenario, noted as DRN, we train the depth branch directly for regression, {\ie} without pre-training a depth classifier. We see that for both cap 50m and 80m, DC-DRN and DRN are on par. However, we obtain the best performance when we introduce the semantic segmentation task during training. Without the semantic information, our DC-DRN and DRN do not yield comparable performance. This suggests that our approach can exploit the additional information provided by semantic information to learn a better depth estimator.

\newcommand{\commentA}{}

\sisetup{detect-weight,mode=text}
\renewrobustcmd{\bfseries}{\fontseries{b}\selectfont}
\renewrobustcmd{\boldmath}{}
\newrobustcmd{\B}{\bfseries}
\newrobustcmd{\IL}{\itshape}
\addtolength{\tabcolsep}{-4.1pt}

\begin{table*}
  \centering
\begin{tabular}{|l||*{11}{c|}}\hline
		& & \multicolumn{5}{c}{ Lower the better}  \vline & \multicolumn{3}{c}{ Higher the better} \vline \\ \hline
		\backslashbox[38mm]{Approaches}{ metrics}&\makebox[3.5em]{cap (m)} &\makebox[3.5em]{rel}&\makebox[3.5em]{sq-rel}&\makebox[3.5em]{rms}&\makebox[3.5em]{rms-log}&\makebox[3.5em]{$\log_{10}$}&\makebox[3.5em]{$\delta \textless  1.25$}&\makebox[3.5em]{$\delta \textless  1.25^2$}&\makebox[3.5em]{$\delta \textless  1.25^3$}\\\hline \hline
		%
		Liu fine-tune \cite{Liu:2016}	& 80   & 0.217	& 1.841   & 6.986	& 0.289	   & -	& 0.647	& 0.882	& 0.961 \\ \hline
		Godard -- K	  \cite{Godard:2017}      & 80	& 0.155	& 1.667	& 5.581	& 0.265	& 0.066	& 0.798	& 0.920	& 0.964  \\ \hline
		Godard -- K	+ CS \cite{Godard:2017} & 80	& 0.124	& 1.24	& 5.393	& 0.230	& \IL 0.052	& 0.855	& 0.946	& 0.975 \\ \hline
		Cao	\cite{Cao:2017} & 80	& 0.115	& -	& 4.712	& 0.198	& -	& \B 0.887	& \IL 0.963	&  0.982 \\ \hline
		
		kuznietsov \cite{Kuznietsov:2017} &	80	& \IL 0.113	& \IL 0.741	& \IL 4.621	& \IL 0.189	& -	& 0.862	& 0.960	& \IL 0.986 \\ \hline
		Ours (DRN)   &  80   & 0.112 & 0.701	& 4.424	& 0.188	& 0.0492  & 0.848 & 0.958 & 0.986 \\ \hline
		
		Ours (DC-DRN)    &  80   & 0.110	& 0.698	& 4.529	& 0.187	& 0.0487  & 0.844 & 0.954 & 0.984 \\ \hline
		\B Ours (DSC-DRN)  &  80   & \B 0.100	& \B 0.601	& \B 4.298	& \B 0.174	& \B 0.044  & \IL 0.874 & \B 0.966 & \B 0.989 \\ \hline

		
		&&&&&&&&& \\ \hline

		
		Garg \cite{Garg:2016} &	50	& 0.169	& 1.512	& 5.763	& 0.236	& -	& 0.836	& 0.935	&0.968 \\ \hline
		Godard -- K	 \cite{Godard:2017}        & 50	& 0.149	& 1.235	& 4.823	& 0.259	& 0.065	& 0.800	& 0.923	& 0.966 \\ \hline
		Godard -- K + CS \cite{Godard:2017}	& 50 	& 0.117	& 0.866	& 4.063	& 0.221	& \IL 0.052	& 0.855	& 0.946	& 0.975 \\ \hline
        Cao	\cite{Cao:2017} & 50	& \IL 0.107	& -	&  3.605	& 0.187	& -	& \B 0.898	& \IL 0.966	& 0.984 \\ \hline
		kuznietsov \cite{Kuznietsov:2017} &	50	& 0.108	& \IL 0.595	& \IL 3.518	& \IL 0.179	& -	& 0.875	& 0.964	& \IL 0.988 \\ \hline
		

		Ours (DRN)     &  50   & 0.109 & 0.618	& 3.702	& 0.182	& 0.0477  & 0.862 & 0.963 & 0.987 \\ \hline
		Ours (DC-DRN)    &  50   & 0.107	& 0.602	& 3.727	& 0.181	& 0.0470  & 0.865 & 0.963 & 0.988 \\ \hline
        
        \B Ours (DSC-DRN)   &  50   & \B 0.096	& \B 0.482	& \B 3.338	& \B 0.166	& \B 0.042   & \IL 0.886 & \B 0.980 & \B 0.995 \\ \hline

\end{tabular}
 \caption{Results on Eigen {\etal}'s KITTI split. DRN - Depth regression network, DC-DRN - Depth regression model with pretrained classification network, DSC-DRN - Depth regression network trained with the conditional flow approach. Evaluation metrics as follows, rel: avg. relative error, sq-rel: square avg. relative error, rms: root mean square error, rms-log: root mean square log error, $log_{10}$: avg. $log_{10}$ error, $\delta<\tau$: $\%$ of pixels with relative error $<\tau$ ($\delta\geq1$; $\delta=1$ no error). Godard -- K means using KITTI for training, and "+ CS " adding Cityscapes too. Bold stands for {\bf best}, italics for \emph{second best}. }
 \label{tab:SOTA_KITTI} 
 \end{table*}

\begin{figure*}
	\centering
	\includegraphics[width=\textwidth]{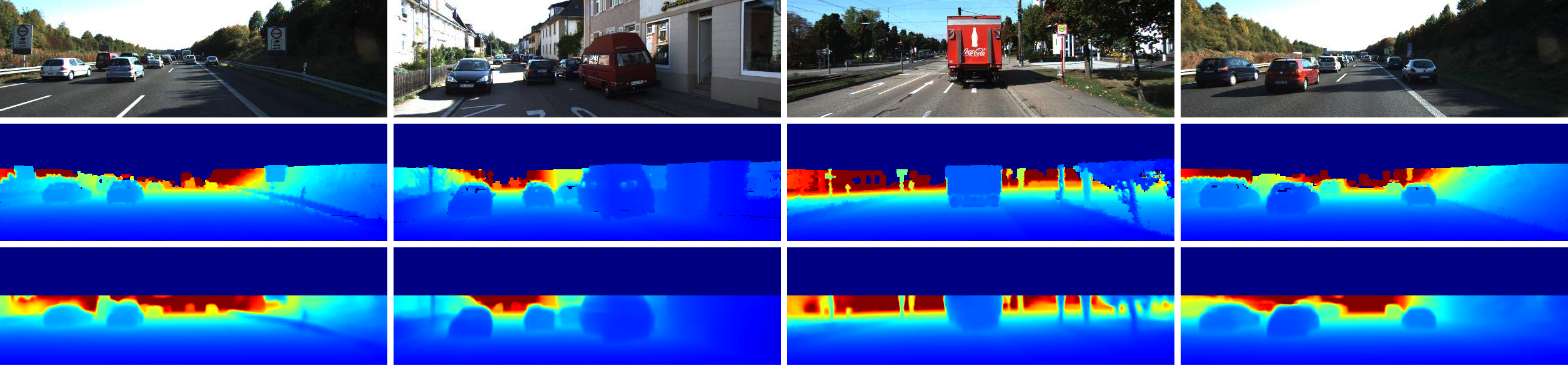}
	\includegraphics[width=\textwidth]{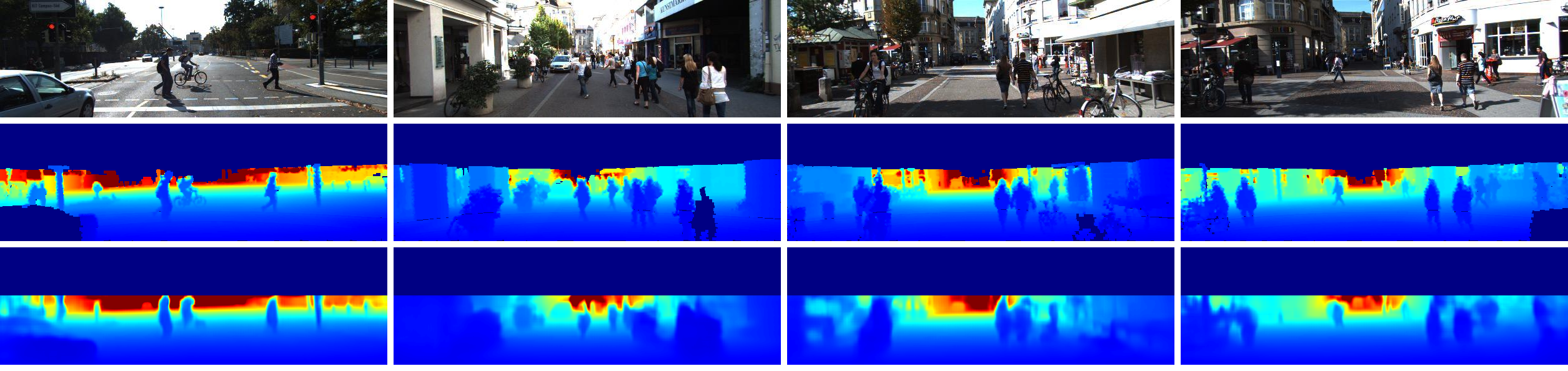}
	\caption{Top to bottom, twice: RGB image (KITTI); depth ground truth; our depth estimation.}
	\label{fig:App_concat_4_J}
\end{figure*}

\begin{figure*}
	
	\centering
	\includegraphics[width=\textwidth]{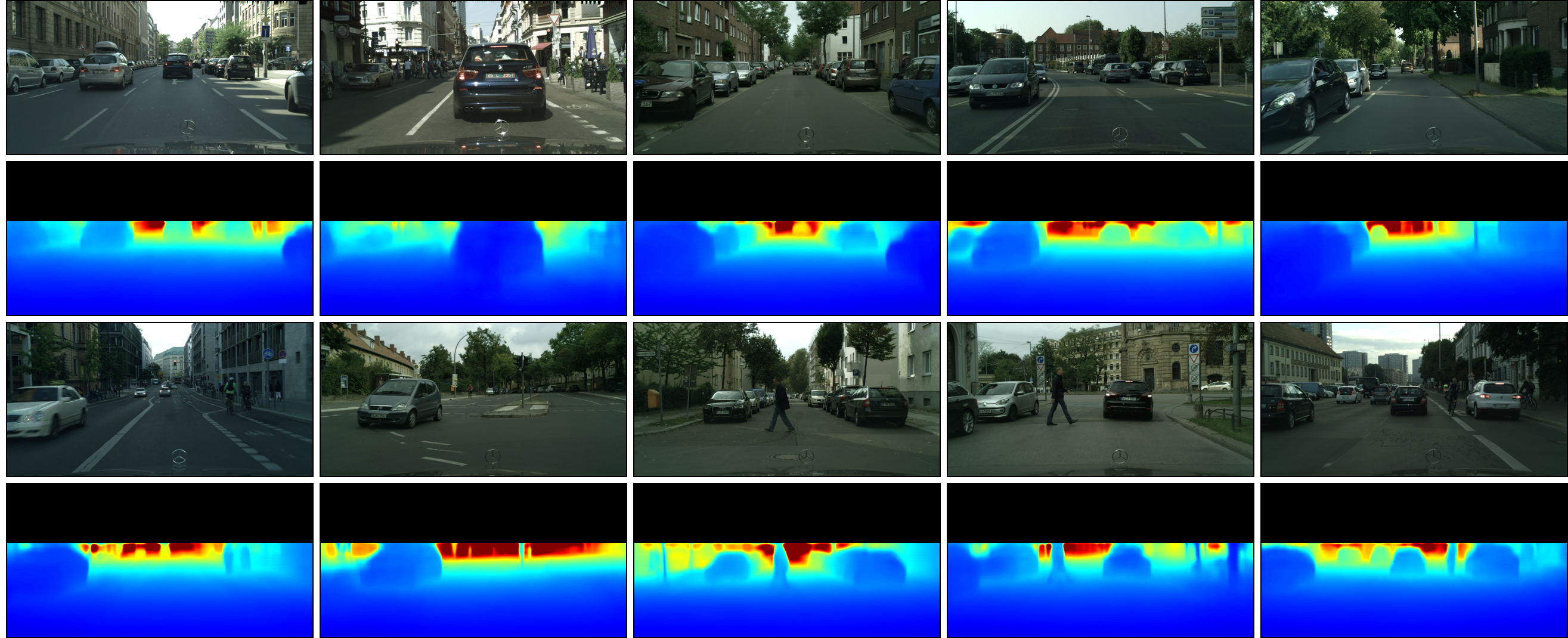}
	\caption{Depth estimation on Cityscapes images not used during training.}
	\label{fig:cityscape_kitti_model}
\end{figure*}

\Fig{App_concat_4_J} shows qualitative results on KITTI. Note how well relative depth is estimated, also how clear are seen vehicles, pedestrians, trees, poles and fences. \Fig{cityscape_kitti_model} shows similar results for Cityscapes; illustrating generalization since the model was trained on KITTI. In this case, images are resized at testing time to KITTI image size ($188\times620$) and the result is resized back to Cityscapes image size ($256\times 512$) using bilinear interpolation.

\section{Conclusion}
\label{sec:conclusion}

We have presented a method to leverage depth and semantic ground truth from different datasets for training a CNN-based depth-from-mono estimation model. Thus, up to the best of our knowledge, allowing for the first time to address outdoor driving scenarios with such a training paradigm ({\ie} depth and semantics). In order to validate our approach, we have trained a CNN using depth ground truth coming from KITTI dataset as well as pixel-wise ground truth of semantic classes coming from Cityscapes dataset. Quantitative results on standard metrics show that the proposed approach improves performance, even yielding new state-of-the-art results. As future work we plan to incorporate temporal coherence in line with works such as \cite{Zhou:2017}.

\section*{ACKNOWLEDGMENT}
Antonio M. L\'opez wants to acknowledge the Spanish project TIN2017-88709-R (Ministerio de Economia, Industria y Competitividad) and the Spanish DGT project SPIP2017-02237, the Generalitat de Catalunya CERCA Program and its ACCIO agency.

\bibliographystyle{IEEEtran}
\bibliography{IEEEabrv,root}

\end{document}